\documentclass[letterpaper, 10 pt, conference]{ieeeconf}  

\IEEEoverridecommandlockouts                              

\usepackage{graphics} 
\usepackage{epsfig} 
\usepackage{mathptmx} 
\usepackage{times} 
\usepackage{amsmath} 
\usepackage{amssymb}  
\usepackage[nolist,nohyperlinks]{acronym}
\usepackage[hidelinks]{hyperref}

\usepackage{bm}
\usepackage{caption} 
\usepackage{subcaption}
\usepackage{graphicx}
\usepackage{makecell}
\usepackage{algorithm}
\usepackage{algpseudocode}
\usepackage{wrapfig}
\usepackage[nolist,nohyperlinks]{acronym}
\usepackage{units}
\usepackage{mathtools}
\usepackage{multirow}
\usepackage{microtype}
\usepackage{xcolor}

\usepackage{gensymb}
\usepackage{balance}
\usepackage{cite}

\usepackage[super]{nth}

\title{\LARGE \bf
Leveraging Neural Radiance Fields for Uncertainty-Aware Visual Localization
}

\author{Le Chen$^{1}$,  Weirong Chen$^{2}$, Rui Wang$^{3}$, Marc Pollefeys$^{3}$
\thanks{$^1$Author is with the Empirical Inference department, Max Planck Institute for Intelligent Systems, Germany, but the work was done while being a member of $^3$. \tt{le.chen@tuebingen.mpg.de}}%
\thanks{$^2$Author is with ETH Zurich, Switzerland, but the work was done while being a member of $^3$. \tt{weirchen@ethz.ch}}
\thanks{$^3$Authors are with the Mixed Reality \& AI Lab, Microsoft, Switzerland. \tt{\{wangr, mapoll\}@microsoft.com}}%
}

\usepackage{xspace}

\makeatletter
\DeclareRobustCommand\onedot{\futurelet\@let@token\@onedot}
\def\@onedot{\ifx\@let@token.\else.\null\fi\xspace}

 \def\Eg{\emph{E.g}\onedot}
\def\ie{\emph{i.e}\onedot}

\def\etal{\emph{et al}\onedot}
\makeatother

\begin{document}

\maketitle

\thispagestyle{empty}
\pagestyle{empty}

\begin{abstract}

As a promising fashion for visual localization, scene coordinate regression (SCR) has seen tremendous progress in the past decade. Most recent methods usually adopt neural networks to learn the mapping from image pixels to 3D scene coordinates, which requires a vast amount of annotated training data. We propose to leverage Neural Radiance Fields (NeRF) to generate training samples for SCR. Despite NeRF's efficiency in rendering, many of the rendered data are polluted by artifacts or only contain minimal information gain, which can hinder the regression accuracy or bring unnecessary computational costs with redundant data. These challenges are addressed in three folds in this paper: (1) A NeRF is designed to separately predict uncertainties for the rendered color and depth images, which reveal data reliability at the pixel level. (2) SCR is formulated as deep evidential learning with epistemic uncertainty, which is used to evaluate information gain and scene coordinate quality. (3) Based on the three arts of uncertainties, a novel view selection policy is formed that significantly improves data efficiency. Experiments on public datasets demonstrate that our method could select the samples that bring the most information gain and promote the performance with the highest efficiency.
%

\end{abstract}

\section{INTRODUCTION}
\label{sec:introduction}

Visual localization, which addresses the problem of estimating the camera pose of a query image in a known environment, is a key component in many robotics applications. 
One way to tackle it is through correspondences between image pixels and 3D map points, which can be obtained by Structure from Motion (SfM) and matching the sparse landmarks to image features. Another popular option is scene coordinate regression (SCR) which directly predicts pixelwise scene coordinates. Empowered by deep learning, recent methods of this category have achieved state-of-the-art performances in small or medium scale scenes~\cite{li2020hierarchical,huang2021vs,tang2021learning}. 
Nevertheless, major challenges remain: (1) obtaining 2D-3D ground truth with sufficient diversity is still computationally and economically expensive in practice, especially using a robot with limited battery or storage; (2) learning an accurate and thorough mapping can be inefficient as it requires training with large amounts of samples for each independent scene.  

Recently, Neural Radiance Fields (NeRF)~\cite{mildenhall2021nerf} has emerged as a powerful paradigm for scene representation. The learned implicit function expresses a compact scene representation and enables realistic view synthesis through differentiable volume rendering.
%
We therefore see a great potential in using NeRF to enhance the data diversity for SCR through synthetic data augmentation.
One critical limitation, though, remains to be addressed: the rendered images usually contain artifacts that may mislead or confuse the network in training. Besides, when rendered randomly, the data may contain large redundancy. 

\begin{figure}[t]
    \setlength{\abovecaptionskip}{0.05cm}
    \setlength{\belowcaptionskip}{-0.7cm}
    \centering
    \includegraphics[width=0.46\textwidth]{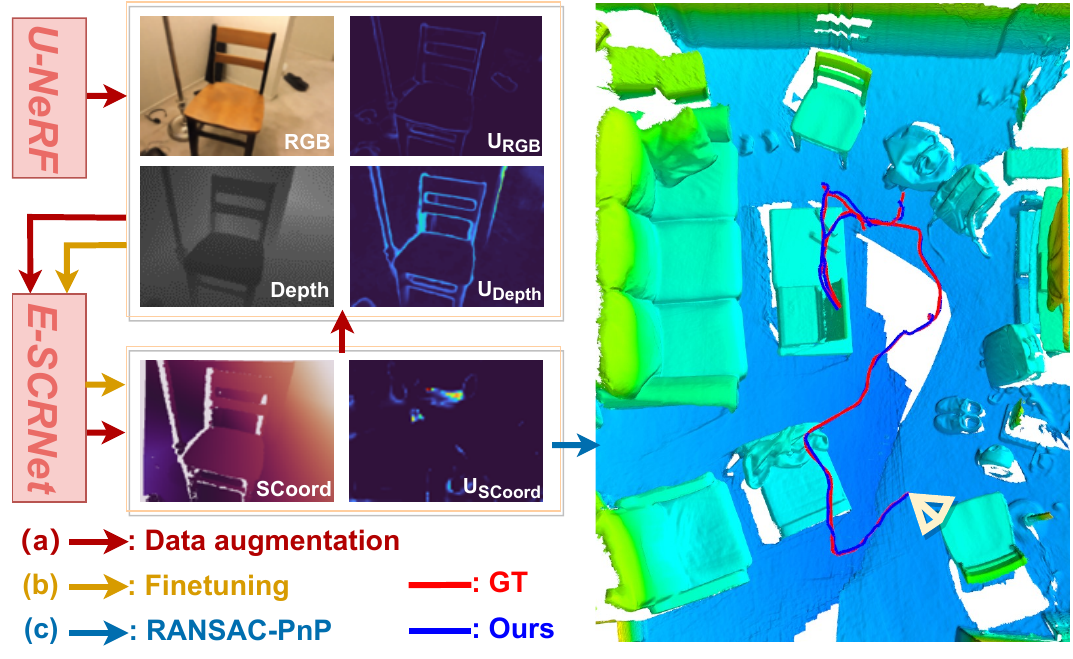}
    \caption[teaser]{\textbf{NeRF-enhanced SCR with uncertainty awareness.} 
    (a) We first train U-NeRF and E-SCRNet to perform informative data augmentation based on the uncertainties of rendered RGB-D images and predicted scene coordinates. (b) The E-SCRNet is finetuned on the augmented dataset considering pixelwise reliability. (c) Finally, the predicted 2D-3D correspondences with uncertainty are fed into PnP to estimate the camera poses.}
    \label{fig:teaser}
\end{figure}

In this paper, we present a novel pipeline that leverages NeRF to address the aforementioned challenges in learning SCR, as shown in Fig.~\ref{fig:teaser}. We train an uncertainty-aware NeRF (U-NeRF) to render RGB-D data with color and depth uncertainties. The rendered data are used to train the SCR network, where the uncertainties are first used to filter out poorly rendered images and then weigh pixels in the regression learning losses. Scene coordinate regression is formulated as evidential deep learning to model the uncertainty of the predicted 3D coordinates. We then present an uncertainty-guided novel view selection policy that selects samples with the most information gain and promotes the network performance with the highest efficiency.
As a result, our method requires only a small portion of the training set but delivers comparable or even better performances than its counterpart trained on the full set.
Our method is the first attempt to separately model color and depth uncertainties for NeRF and utilize them for learning SCR. Also for the first time, we formulate SCR as evidential deep learning and propose an uncertainty-guided policy to filter the rendered data for model evidence. Our method is orthogonal to SCR networks and can serve as a plug-and-play module.

\label{sec:introduction}
\section{RELATED WORKS}
\label{sec:relatedwork}

\noindent\textbf{Visual localization} methods can be roughly categorized into three groups: Direct pose regression using Convolutional Neural Networks (CNN)~~\cite{kendall2017geometric,kendall2015posenet,sattler2019understanding,walch2017image,chen2021direct,chen2022dfnet};
Image retrieval with reference images tagged with known poses~\cite{sattler2012image,arandjelovic2013all,torii201524,arandjelovic2016netvlad};
Pose estimation from 2D-3D correspondences which are usually obtained by sparse feature matching~\cite{dusmanu2019d2,sarlin2019coarse,detone2018superpoint,zhang2018learning,zhou2016evaluating,shotton2013scene,brachmann2021dsacstar,sattler2016efficient,sattler2012improving,taira2018inloc, sarlin2021back, lindenberger2021pixel}.
Scene coordinate regression adds another important branch to the third category. 
It obtains the correspondences by regressing dense 3D scene coordinates for the query image using random forests or neural networks, then calculates the final camera pose via RANSAC-PnP. 
Shotton \etal~\cite{shotton2013scene} propose to regress scene coordinates using a Random Forest, followed by several variants~\cite{cavallari2017fly,meng2017backtracking,meng2018exploiting,valentin2015exploiting,valentin2016learning}. DSAC~\cite{brachmann2017dsac} and its follow-ups~\cite{brachmann2018learning,brachmann2021dsacstar} employ CNNs to predict the coordinate map and propose a differentiable approximation of RANSAC for end-to-end training. Li \etal~\cite{li2020hierarchical} partition the scene into regions and hierarchically predict scene coordinates by adding several classification layers into a regression network. 
Huang \etal~\cite{huang2021vs} partition the 3D surfaces into 3D patches and train a segmentation network to obtain correspondences between image pixels and patch centers. As a main bottleneck, these methods require large amounts of posed images to train their models. 

\vspace{0.1cm}
\noindent\textbf{NeRF}~\cite{mildenhall2021nerf} was introduced as a powerful technique for synthesizing novel views of complex scenes. 
Follow-up works try to add depth supervision to enhance quality~\cite{roessle2022dense,wei2021nerfingmvs,deng2022depth}, reduce data requirement~\cite{chen2021mvsnerf,yu2021pixelnerf,xu2022sinnerf}, handle noisy or unknown camera pose~\cite{wang2021nerf,lin2021barf}, speed up optimization or rendering~\cite{Chen2022ECCV,sun2022direct,garbin2021fastnerf,reiser2021kilonerf,mueller2022instant} and extend to large-scale environments~\cite{tancik2022block,rematas2022urban,turki2022mega}. NeRF has been widely adopted for applications like navigation~\cite{adamkiewicz2022vision}, manipulation~\cite{IchnowskiAvigal2021DexNeRF}, active 3D object reconstuction~\cite{lee2022uncertainty}, data augmentation for learning object descriptors~\cite{yen2022nerf}. 
Sharing our spirit of adopting NeRF to synthesize novel views for visual localization, 
Zhang \etal iteratively refine camera poses based on feature matching between rendered and real images~\cite{zhang2021reference}. iNeRF~\cite{yen2021inerf} leverages NeRF's differentiability and estimates camera poses in an analysis-by-synthesis manner. Chen \etal combine NeRF with pose regression by directly comparing the query and the rendered image~\cite{chen2021direct}. The work is extended by adding feature matching with a random view synthesis strategy~\cite{chen2022dfnet}. LENS~\cite{moreau2022lens} deploys NeRF-W~\cite{martin2021nerf} to synthesize views uniformly within the scene boundary to train a pose regression network. 
In this work, we propose to use NeRF to render data with high information gain for learning SCR. 

\begin{figure*}[ht]
    \centering
    \setlength{\abovecaptionskip}{0cm}
    \setlength{\belowcaptionskip}{-0.65cm}
    \includegraphics[width=0.85\textwidth]{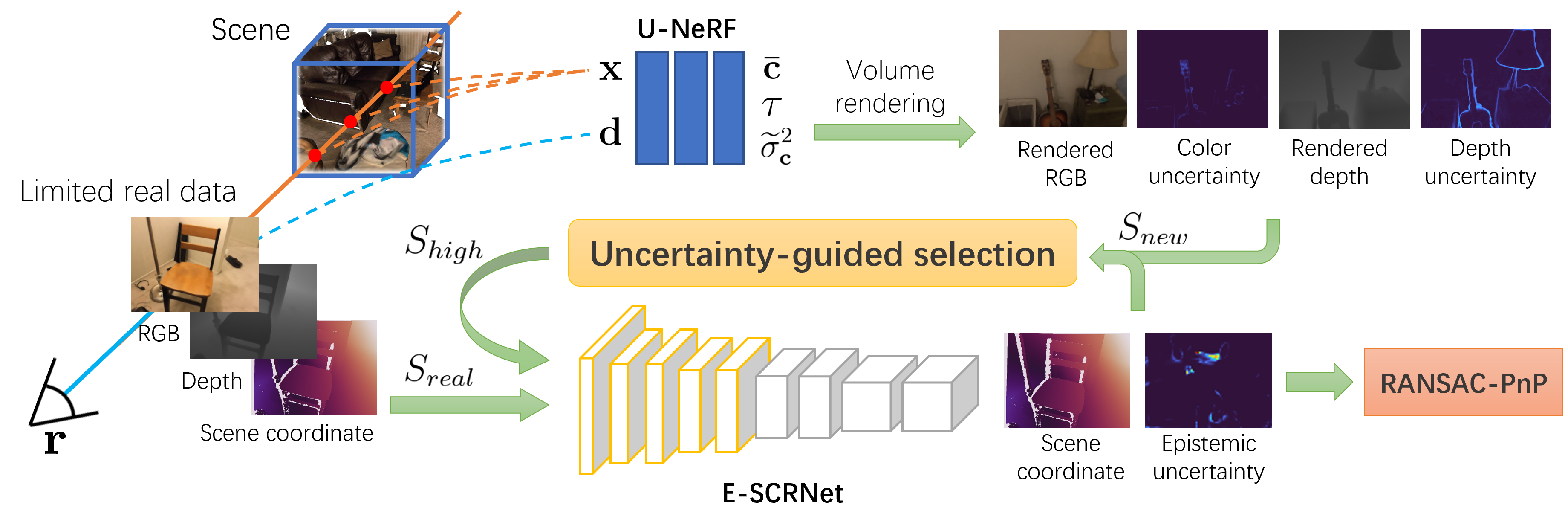}\caption[overview]{Overview. We first train an uncertainty-aware NeRF model and the evidential scene coordinate regression network (E-SCRNet) with the available data. We then render novel views with corresponding color and depth uncertainty maps. We apply an uncertainty-guided policy to select new views and gather them with the available data to finetune the E-SCRNet.}
    \label{fig:overview}
\end{figure*}

\noindent\textbf{Uncertainty estimation} for neural networks is relevant to assessing confidence, information gain and capturing outliers~\cite{kendall2017uncertainties,georgakis2021learning,georgakis2022uncertainty}. 
Bayesian neural networks~\cite{mackay1995bayesian,kononenko1989bayesian,kendall2017uncertainties} learn the posterior distribution of network weights and estimate it using variational inference, which can be approximated by Dropout or Deep Ensembles~\cite{gal2016dropout,kingma2015variational,lakshminarayanan2017simple,jain2020maximizing}.
However, they require expensive sampling during inference.
Recently, evidential theory is incorporated into neural networks~\cite{sensoy2018evidential,amini2020deep}, where training data add support to a learned higher-order evidential distribution.
By learning the distribution hyperparameters, uncertainty can be estimated by a single forward pass, circumventing sampling during inference.
For NeRF, NeRF-W~\cite{martin2021nerf} integrates uncertainty to attenuate transient scene elements. S-NeRF~\cite{shen2021stochastic} encodes the posterior distribution over all the possible radiance fields and estimates uncertainty by sampling.
CF-NeRF~\cite{shen2022conditional} follows a similar strategy and learns the radiance field distributions by coupling Latent Variable Modelling and Conditional Normalizing Flows. ActiveNeRF~\cite{pan2022activenerf} models scene point radiance as Gaussian and uses uncertainty as a criterion for capturing new inputs. 

\section{METHOD}
\label{sec:method}
SCR methods rely on large amounts of annotated training data that are hard to obtain in practice. 
To leverage synthetic data, NeRF can be used to render samples for arbitrary viewpoints using just a few posed images.
However, the images rendered may be of low quality or highly redundant, which can lower the SCR network's accuracy and efficiency.
This motivates us to assess the quality and information gain of each view to make the best use of the synthetic data.

 We first design U-NeRF and E-SCRNet. 
 U-NeRF predicts color and depth uncertainties, reflecting the image quality, while E-SCRNet estimates the epistemic uncertainty of the predicted scene coordinates.
 Given a small set of real posed RGB-D images $S_{real}$, we train U-NeRF and an initial version of E-SCRNet. U-NeRF is used to generate a novel view dataset $S_{new}$ by synthesizing more views within the scene boundary  (Sec.~\ref{sec:view_synthesis}). We feed $S_{new}$ to E-SCRNet to get the epistemic uncertainties, which reflect the information gain of each sample (Sec.~\ref{sec:sc_uncertainty}). 
 We prune invalid views from $S_{new}$ based on U-NeRF's uncertainties, then select views with high information gain $S_{high}$. The final augmented dataset, $S_{aug} = S_{real} \cup S_{high}$, is used to finetune E-SCRNet, increasing localization performance with higher data efficiency (Sec.~\ref{sec:novel_views_selection}). Our overall pipeline is illustrated in Fig.~\ref{fig:overview}.

\subsection{View synthesis with uncertainty estimation}
\label{sec:view_synthesis}

\noindent\textbf{Preliminaries on NeRF.}
NeRF models the scene as a continuous function using a multilayer perceptron (MLP). For a 3D position $\mathbf{x}$ and a viewing direction $\mathbf{d}$ that are transformed using positional encoding $\phi(\text{·})$, the learned implicit function outputs a volume density $\tau$ and a view-dependent RGB color $\mathbf{c}$.
To obtain the color of a pixel, consider the ray $\mathbf{r}(t) = \mathbf{o} + t\mathbf{d}$ emanating from the camera center $\mathbf{o} \in \mathbb{R}^3$ and traversing the range $[t_n, t_f]$. Volume rendering computes light radiance by integrating the radiance along the ray. NeRF approximates it using hierarchical stratified sampling~\cite{max1995optical} by partitioning [$t_n, t_f$] into $N$ bins and sampling uniformly in each bin.
The expected color $\hat{\mathbf{C}}(\mathbf{r})$ can be approximated by
\begin{equation}
    \hat{\mathbf{C}}(\mathbf{r}) = \sum_{i=1}^{N} T_i \ (1-\exp{(-\tau_i \delta_i)}) \ \mathbf{c}_i = \sum_{i=1}^{N} w_i \mathbf{c}_i,
\label{eq:quadrature}
\end{equation}
where $T_i = \exp{(-\sum_{j=1}^{i-1} \tau_j \delta_j)}$ and $\delta_i = t_{i+1} - t_i$. $\hat{\mathbf{C}}(\mathbf{r})$ therefore is a weighted sum of the color samples $\mathbf{c}_i$.

We aim to use NeRF to synthesize novel views for training the SCR network. However, the synthesized views may contain noises, blur, and other artifacts caused by varied imaging conditions of inputs. Training with such noisy samples can mislead the network from clean distribution and decrease its performance. To reason about the reliability of the rendered data, we formulate our scene representation with uncertainty estimation, named U-NeRF. One key observation is that the uncertainties of the rendered color and depth images should be modeled separately, as they present quite different distributions as shown in Fig.~\ref{fig:uncertainty}.

\begin{figure}[t]
    \setlength{\abovecaptionskip}{0.1cm}
    \setlength{\belowcaptionskip}{-0.8cm}
    \centering
    \includegraphics[width=0.32\textwidth]{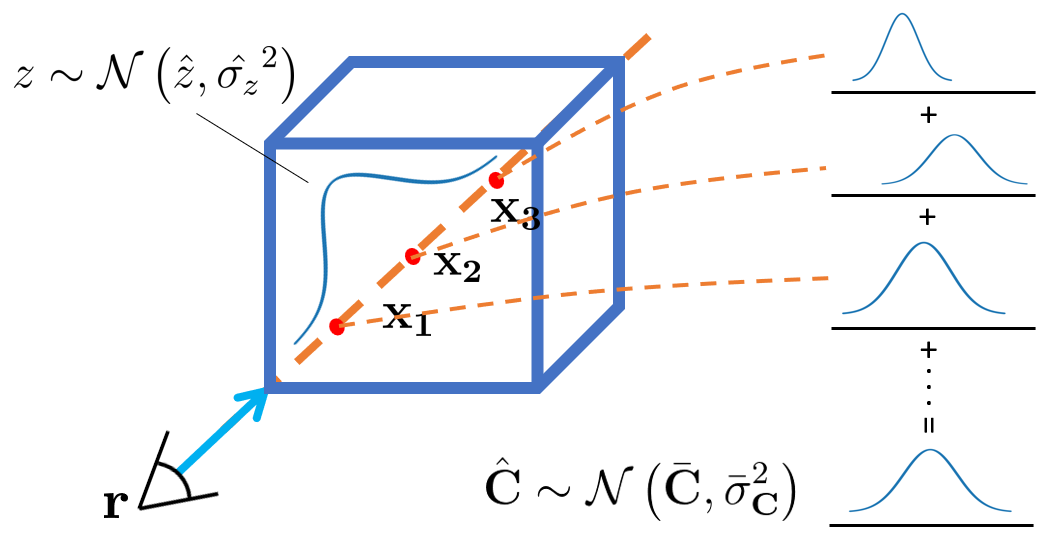}\caption[nerf uncertainty]{U-NeRF. If the distribution of a scene point's radiance is Gaussian, the rendered pixel color is the weighted sum of Gaussians, thus still a Gaussian. We also assume the ray termination distribution $z$ to be Gaussian.}
    \label{fig:nerf_uncertainty}
\end{figure}

\noindent\textbf{Color uncertainty.}
To estimate the color uncertainty, we assume the radiance value of a scene point to be Gaussian $\mathbf{c}_i \sim \mathcal{N}\left(\bar{\mathbf{c}}_i, \bar{\sigma}_{\mathbf{c}_i}^2\right)$. The mean $\bar{\mathbf{c}}_i$ is the predicted radiance and the variance $\bar{\sigma}_{\mathbf{c}_i}^2$ captures the color uncertainty of a certain scene point. Employing Bayesian learning~\cite{kendall2017uncertainties,martin2021nerf,pan2022activenerf}, we add an additional branch to predict $\bar{\sigma}_{\mathbf{c}_i}^2$ after injecting viewing directions: $(\tau, \bar{\mathbf{c}}, \widetilde{\sigma}_\mathbf{c}^2) = \text{MLP}\left(\phi\left(\mathbf{x},\mathbf{d}\right)\right)$.
$\widetilde{\sigma}_\mathbf{c}^2$ is processed by a Softplus to obtain valid variance $\bar{\sigma}_\mathbf{c}^2$.
When performing volume rendering as in Eq.\ref{eq:quadrature}, the rendered pixel color can be viewed as a weighted sum of radiance colors $\mathbf{c}_i$.
%
%
%
By assuming that the distributions of different sampled scene points on one ray are independent, and the distributions of sampled rays are independent, we can derive that the rendered pixel color follows a Gaussian distribution:
$
\hat{\mathbf{C}} \sim \mathcal{N}\left(\bar{\mathbf{C}}, \bar{\sigma}_\mathbf{C}^2\right) \sim \mathcal{N}\left(\sum_{i=1}^{N} w_i \bar{\mathbf{c}}_i, \sum_{i=1}^{N} w_i^2 \bar{\sigma}_{\mathbf{c}_i}^2\right) ,
$
where $w_i$ is defined in Eq.\ref{eq:quadrature}, $N$ is the number of sampled points along the ray. $\bar{\mathbf{C}}$ and $\bar{\sigma}_\mathbf{C}^2$ are the mean and variance of the rendered pixel color (see Fig.~\ref{fig:nerf_uncertainty}). 
From a maximum likelihood perspective, we can optimize the model by minimizing the negative log-likelihood (NLL) loss for sampled rays $\mathbf{r}$~\cite{martin2021nerf,pan2022activenerf}:
\begin{equation}
\mathcal{L}_{\mathrm{NLL}}= \sum_{\mathbf{r} \in R}\underbrace{\left(\frac{\|\mathbf{C}\left(\mathbf{r}\right)-\hat{\mathbf{C}}\left(\mathbf{r}\right)\|_2^2}{\hat{\sigma}_\mathbf{C}^2\left(\mathbf{r}\right)}+\log \hat{\sigma}_\mathbf{C}^2\left(\mathbf{r}\right)\right)}_{\mathcal{L}(\mathbf{r})}.
\end{equation}
However, it leads to sub-optimal mean fits and premature convergence as badly-fit regions receive increasingly less weights. To address the ignorance of hard-to-fit regions, we add a variance-weighting term $\sigma_\mathbf{C}^{2\zeta}$ that acts as a factor on the gradient~\cite{Seitzer2022PitfallsOfUncertainty} as
$
\mathcal{L}_{\mathrm{color}}=\sum_{\mathbf{r} \in R}\left\lfloor\hat{\sigma}_\mathbf{C}^{2\zeta}\right\rfloor \mathcal{L}(\mathbf{r}),
$
where $\left\lfloor\text{·}\right\rfloor$ denotes the stop gradient operation which prevents the gradient from flowing through inside the parenthesis, making the variance-weighting term an adaptive learning rate.
The parameter $\zeta$ interpolates between NLL ($\zeta=0$) and MSE ($\zeta=1$) while providing well-behaved uncertainty estimates.
Note that NeRF-W~\cite{martin2021nerf} applies the volume rendering over $\sigma_\mathbf{C}$ instead of ${\sigma^2_\mathbf{C}}$, which is not theoretically grounded. 
Our approach is similar to ActiveNeRF~\cite{pan2022activenerf}, but we adopt a $\zeta$-NLL loss to cope with sub-optimal performance.

\noindent\textbf{Depth uncertainty.}
Eq.\ref{eq:quadrature} can be slightly modified to obtain the expected depth $\hat{z}(\mathbf{r})$ of each ray and its variance $\hat{\sigma}_z(\mathbf{r})^2$:
%
%
$
\hat{z}(\mathbf{r})=\sum_{i=1}^N w_i t_i, \quad \hat{\sigma}_z(\mathbf{r})^2=\sum_{i=1}^N w_i\left(t_i-\hat{z}(\mathbf{r})\right)^2.
$
\cite{deng2022depth} demonstrates that adding depth supervision improves the reconstruction quality and leads to a change in the weight distribution from multimodal to unimodal. We assume the ray termination distribution to be Gaussian $z_i \sim \mathcal{N}\left(\hat{z}_i, \hat{\sigma}_z(\mathbf{r})^2\right)$ and adopt a NLL on the output depth~\cite{roessle2022dense}:
\begin{equation}
\mathcal{L}_{\mathrm{depth}}= \sum_{\mathbf{r} \in R}\left(\frac{\left\|\hat{z}(\mathbf{r}))-z(\mathbf{r}))\right\|_2^2}{\hat{\sigma}_z(\mathbf{r})^2}+\log \hat{\sigma}_z(\mathbf{r})^2\right),
\end{equation}
where $z(\mathbf{r})$ is the target depth. The variance $\hat{\sigma}_z(\mathbf{r})^2$ captures the uncertainty of the rendered depth (see Fig.~\ref{fig:nerf_uncertainty}).

 \noindent\textbf{Training U-NeRF.}
 We optimize our scene representation with the objective function
\begin{equation}
    \mathcal{L}_{\mathrm{overall}} = \mathcal{L}_{\mathrm{color}} + \lambda \mathcal{L}_{\mathrm{depth}}.
\end{equation}
To reduce the training cost, we adopt a single MLP and optimize with depth-guided sampling~\cite{roessle2022dense}.

\subsection{Evidential scene coordinate regression}
\label{sec:sc_uncertainty}
For evaluating the potential contribution of each synthetic sample, we consider not only the rendering quality (reflected by uncertainties from U-NeRF) but also how informative the sample is for learning SCR. To this end, the epistemic uncertainty~\cite{georgakis2021learning,georgakis2022uncertainty,pathak2019self,seung1992query} can be adopted to reflect the information gain the sample brings to the network. Compared to aleatoric uncertainty which represents the inherent randomness in data that cannot be explained away, epistemic uncertainty captures the uncertainty over network parameters and describes the confidence of the prediction~\cite{kendall2017uncertainties}. Therefore, data samples with high epistemic uncertainty are associated with increased information gain. We propose to formulate scene coordinate regression from the perspective of evidential deep learning~\cite{sensoy2018evidential,amini2020deep}, which enables fast sampling-free estimation of epistemic uncertainty. Since we aim to develop a general pipeline that can be applied to any SCR network, we take the simple regression-only baseline (SCRNet) from~\cite{li2020hierarchical} as an example in this paper.

We assume the observed scene coordinates $q_i$ are drawn $\mathrm{i.i.d.}$ from an underlying Gaussian distribution with unknown mean $\mu_q$ and variance ${\sigma_q}^2$. To estimate the posterior distribution $p\left(\mu_q, {\sigma_q}^2 \mid q_1, \ldots, q_N\right)$, we place priors over the likelihood variables with a Gaussian prior on ${\mu_q} \sim \mathcal{N}\left(\gamma, {\sigma_q}^2 v^{-1}\right)$ and an Inverse-Gamma prior on ${\sigma_q}^2 \sim \Gamma^{-1}(\alpha, \beta)$.
Assuming that the posterior distribution can be factorized, we can approximate it with a Normal Inverse-Gamma (NIG) distribution~\cite{amini2020deep}
Sampling an instance from the NIG distribution yields a Gaussian distribution
from which scene coordinates $q_j$ are drawn. Hence, the NIG distribution can be interpreted as the evidential distribution on top of the unknown likelihood distribution from which observations are drawn. By estimating the NIG hyperparameters $(\gamma, v, \alpha, \beta)$, we can compute the prediction $\mathbb{E}[\mu]=\gamma$ and epistemic uncertainty $\operatorname{Var}[\mu]=\beta / (v(\alpha-1))$.
To train a network to output the correct scene coordinates and the hyperparameters of the corresponding NIG distribution, we modify the last layer of SCRNet to predict $(\gamma, v, \alpha, \beta)$, and maximize the model fit while minimizing evidence on errors~\cite{amini2020deep}:
\vspace{-0.2cm}
\begin{equation}
\begin{aligned}
&\mathcal{L}_{\mathrm{coord}}= \sum_{\mathbf{q} \in S}\left[\left(\alpha+1/2 \right) \log \left(\left(q-\gamma\right)^2 \nu+2\beta\left(1+\nu\right)\right)\right] \\
&+\sum_{\mathbf{q} \in S}\left[\left(1/2 \right)\log \left(\frac{\pi}{\nu}\right)-\alpha \log \left(2\beta\left(1+\nu\right)\right)\right] \\
&+\sum_{\mathbf{q} \in S}\left[\log \left(\frac{\Gamma\left(\alpha\right)}{\Gamma\left(\alpha+1/2 \right)}\right) + \left|q-\gamma\right| \cdot\left(2\nu+\alpha\right)\right].
\end{aligned}
\label{eq:coord}
\end{equation}

In this way, the E-SCRNet regresses the scene coordinates for each pixel with corresponding uncertainty. We consider the epistemic uncertainty as a proxy for information gain and therefore a score in the following novel view selection policy. In addition, the uncertainty also reflects the confidence of the predicted 2D-3D correspondences. We can use it to filter out unreliable correspondences in RANSAC-PnP.

\subsection{Selection policy of novel views}
\label{sec:novel_views_selection}

With color and depth uncertainties from U-NeRF and epistemic uncertainty from E-SCRNet prepared, we are ready to select the most informative samples to boost the localization performance without introducing much computational cost. To generate the initial synthetic training data $S_{new}$, we use U-NeRF to render RGB-D images with uncertainty estimation from different viewpoints within the scene. Note that this view synthesizing process can be very fast with highly efficient NeRF rendering ~\cite{garbin2021fastnerf,mueller2022instant,yu2021plenoctrees}. We then propose the following two steps for data selection, \ie, rendering quality based pruning and information gain based selection.


 \noindent\textbf{Pruning based on rendering quality.}
Given the rendered color and depth uncertainties, we use the following criteria to remove inferior candidates: (1) views with rendered depth smaller than $z_{min}$ are too close to the scene structure and are considered invalid; (2) views with large depth uncertainty carry incorrect geometric information of the scene; (3) views with large color uncertainty contain noisy image texture that would confuse the SCRNet. This first round of pruning improves the overall quality of the rendered novel view set, but $S_{new}$ may still contain unnecessary redundancy. \Eg, given a new view, if the SCRNet has already learned it well from $S_{real}$ and is very confident in its prediction, we know that the new view will bring little information gain. Therefore, another view selection step is performed.

 \noindent\textbf{Selection based on information gain.}
As mentioned in Sec.~\ref{sec:sc_uncertainty}, the epistemic uncertainty can be used as a proxy for information gain. We therefore propose a view selection policy guided by the scene coordinate epistemic uncertainty. For each of the rendered images in $S_{new}$, we apply the E-SCRNet pre-trained from $S_{real}$ to get the epistemic uncertainty map (together with the scene coordinate map). We aim to use the mean epistemic uncertainty as a novel view score. To remove the influence of rendered artifacts, the color and depth uncertainties provided by U-NeRF are used to filter out unreliable pixels. The images with the top-k scores are considered to provide high information gain and are selected to form the novel view set $S_{high}$.
%
%


\begin{figure*}[ht]
\setlength{\abovecaptionskip}{0.1cm}
    \centering
     \begin{subfigure}[b]{0.16\textwidth}
         \centering
         \includegraphics[width=\textwidth]{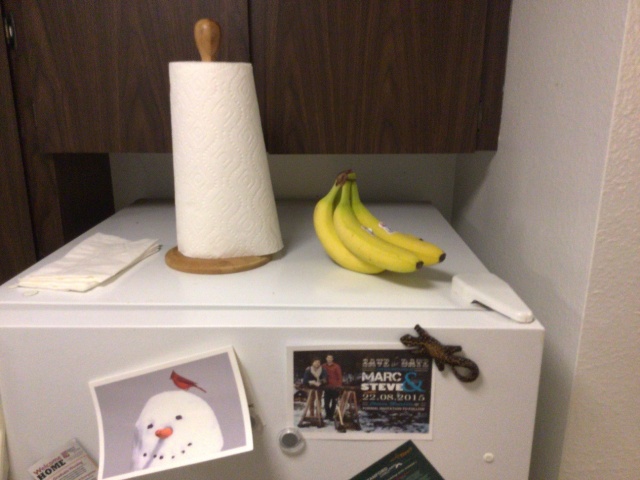}
     \end{subfigure}
     \begin{subfigure}[b]{0.16\textwidth}
         \centering
         \includegraphics[width=\textwidth]{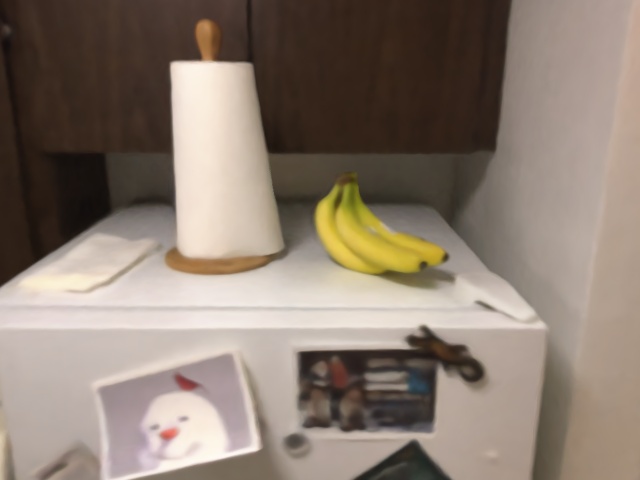}
     \end{subfigure}
     \begin{subfigure}[b]{0.16\textwidth}
         \centering
         \includegraphics[width=\textwidth]{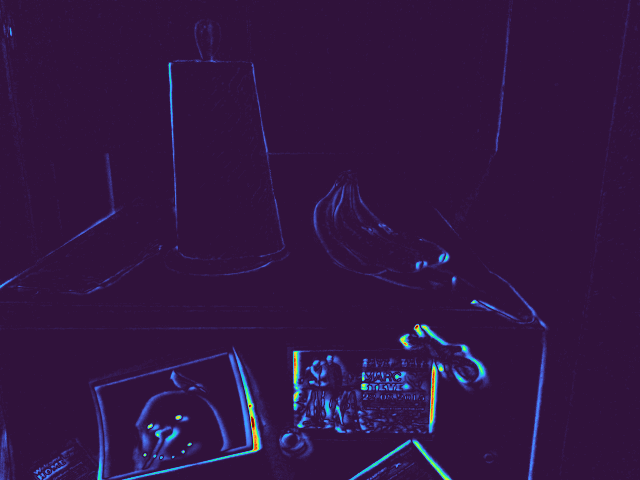}
     \end{subfigure}
    \begin{subfigure}[b]{0.16\textwidth}
         \centering
         \includegraphics[width=\textwidth]{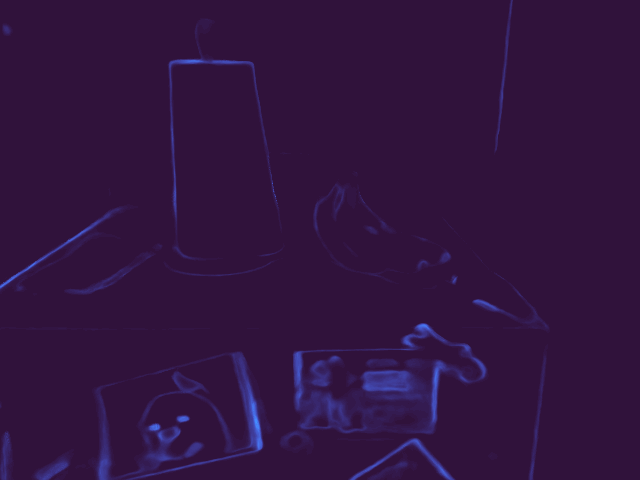} 
     \end{subfigure}
    \begin{subfigure}[b]{0.16\textwidth}
         \centering
         \includegraphics[width=\textwidth]{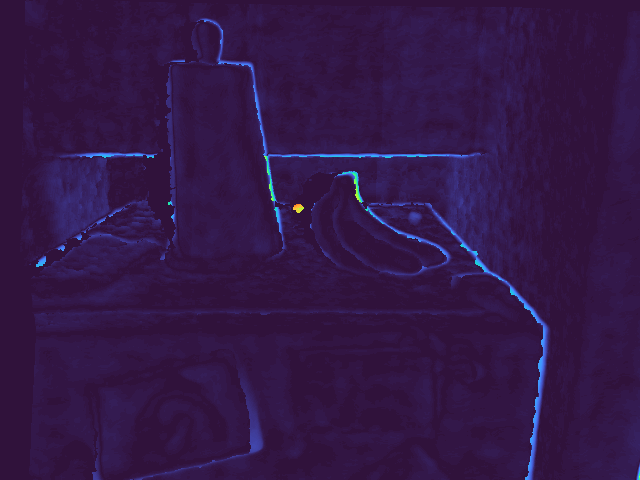}
     \end{subfigure}
    \begin{subfigure}[b]{0.16\textwidth}
         \centering
         \includegraphics[width=\textwidth]{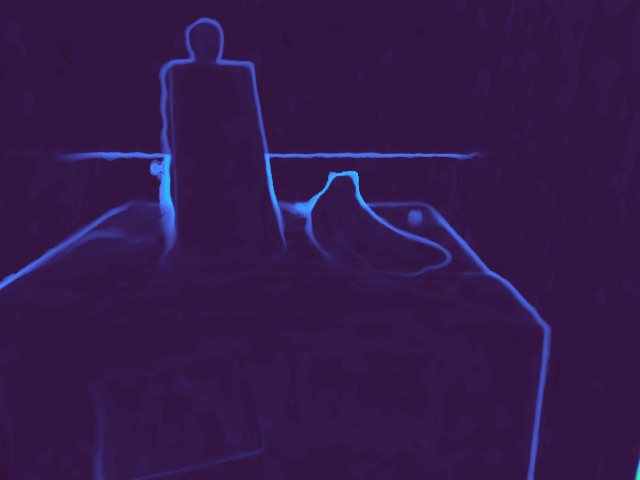}
     \end{subfigure}     
     \\[0.2ex]

     \begin{subfigure}[b]{0.16\textwidth}
         \centering
         \includegraphics[width=\textwidth]{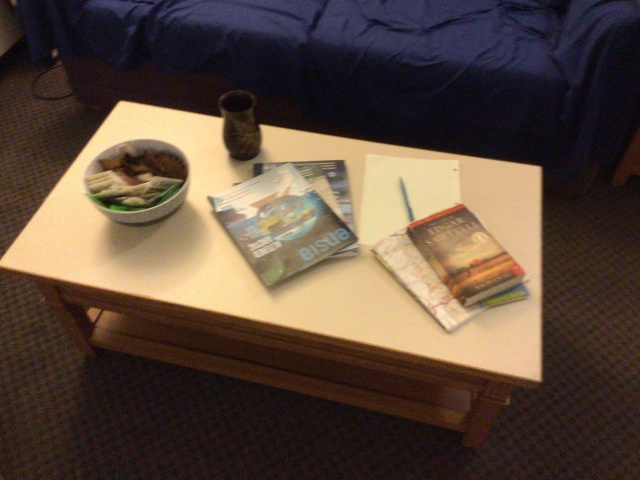}
         \caption{GT color}
     \end{subfigure}
     \begin{subfigure}[b]{0.16\textwidth}
         \centering
         \includegraphics[width=\textwidth]{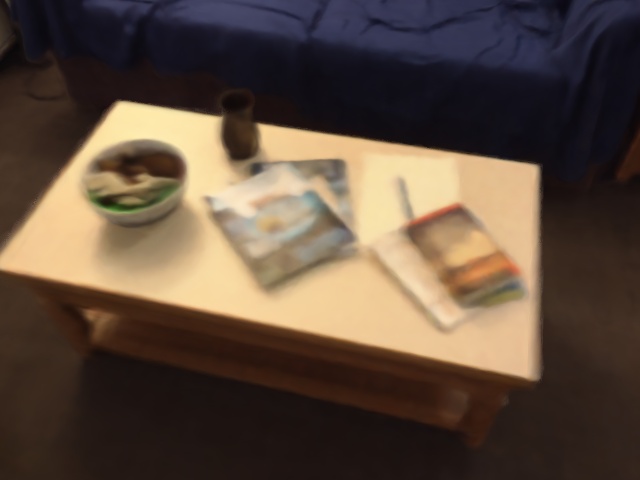}
         \caption{Rendered color}
     \end{subfigure}
     \begin{subfigure}[b]{0.16\textwidth}
         \centering
         \includegraphics[width=\textwidth]{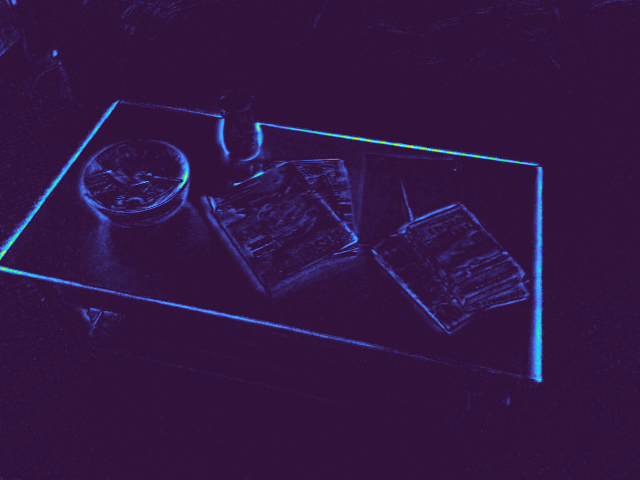}
         \caption{Color error}
     \end{subfigure}
    \begin{subfigure}[b]{0.16\textwidth}
         \centering
         \includegraphics[width=\textwidth]{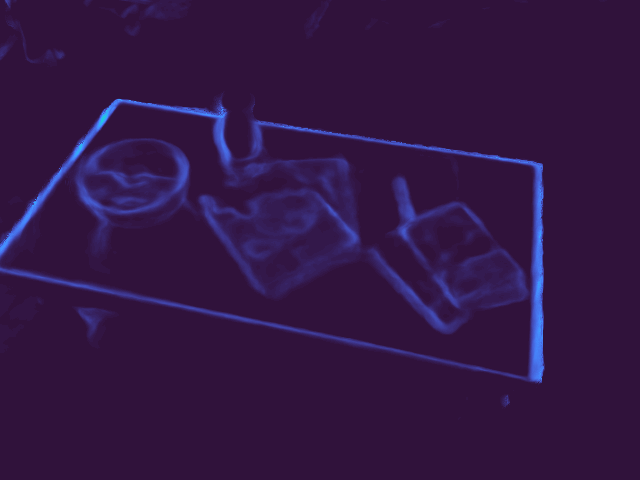}
         \caption{Color uncertainty}
     \end{subfigure}
    \begin{subfigure}[b]{0.16\textwidth}
         \centering
         \includegraphics[width=\textwidth]{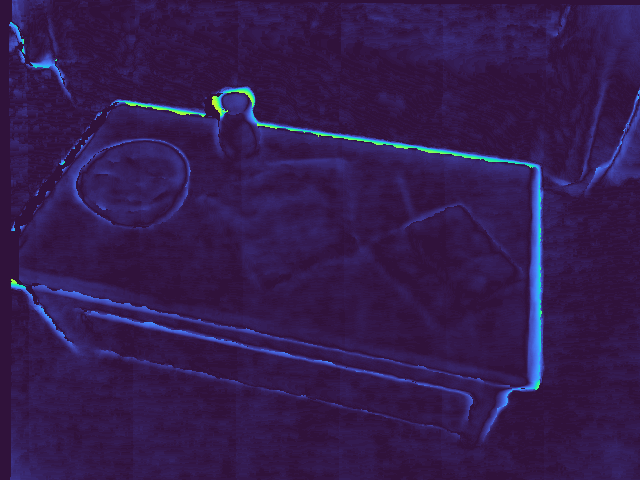}
         \caption{Depth error}
     \end{subfigure}
    \begin{subfigure}[b]{0.16\textwidth}
         \centering
         \includegraphics[width=\textwidth]{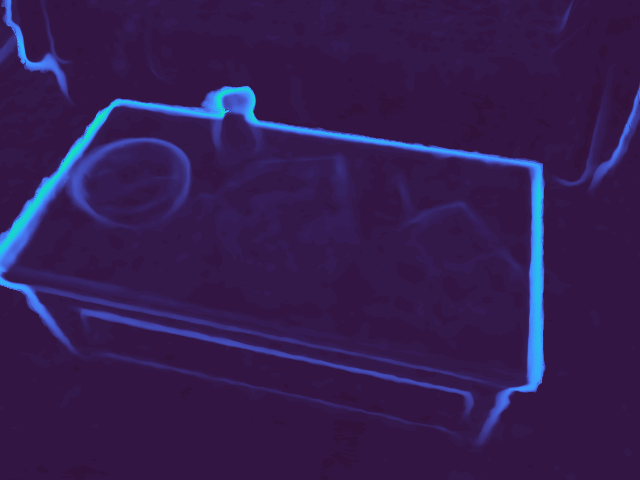}
         \caption{Depth uncertainty}
     \end{subfigure} 
    \caption[uncertainty]{Qualitative results of U-NeRF uncertainty estimation.}
    \label{fig:uncertainty}
\vspace{-0.3cm}
\end{figure*}

 \noindent\textbf{Training with pixelwise uncertainty.}
In addition to bad sample pruning and view selection on image level, RGB-D uncertainties also indicate noises and artifacts on the pixel level. To alleviate the influences of these noisy rendered signals, we further adopt the pixel-wise RGB-D uncertainties into loss design.
Firstly, pixels with uncertainties beyond the predefined thresholds are ignored in the loss. Secondly, they are used to weigh the remaining pixels. Since E-SCRNet is trained with color-depth pairs, we consider both uncertainties and formulate the weighting function as $\kappa=e^{-2({\sigma_c}^2 + {\sigma_z}^2)}$,
then replace $\left(q-\gamma\right)$ with $\kappa\left(q-\gamma\right)$ in Eq.\ref{eq:coord}. This weighting encourages E-SCRNet to memorize well-rendered regions better while paying less attention to uncertain regions.


\section{Experiments}
\label{sec:experiment}

In this section, we conduct experiments to evaluate our pipeline. The results show that U-NeRF and E-SCRNet could provide meaningful uncertainties and our pipeline is able to improve camera pose estimation accuracy by a large margin.

\begin{table*}[h]
\setlength{\abovecaptionskip}{0.1cm}
\centering
\scriptsize
\caption{Visual localization results on Replica. We report median and mean translation and rotation errors (cm,\degree). We also report the sum of the L1 distance between predicted scene coordinates and ground truth in $m$. The best results under the few views training are labeled in bold.}
\begin{tabular}{c|ccc|ccc|cc|ccc}
\hline
          & \multicolumn{3}{c|}{SCRNet~\cite{li2020hierarchical}}             & \multicolumn{3}{c|}{SCRNet-ID~\cite{ng2021reassessing}}                       & \multicolumn{2}{c|}{SRC~\cite{dong2022visual}}            & \multicolumn{3}{c}{Ours}                                                                                     \\ \cline{2-12} 
          & Median$\downarrow$                   & Mean$\downarrow$ & Dist. & Median$\downarrow$                   & Mean$\downarrow$ & Dist.$\downarrow$ & Median$\downarrow$                   & Mean$\downarrow$ & Median$\downarrow$                   & Mean$\downarrow$                     & Dist.$\downarrow$              \\ \hline
room\_0   & 2.05, 0.33                           & 2.38, 0.36       & 0.27  & 2.33, 0.28                           & 2.55, 0.32       & 0.24              & 2.78, 0.54                           & 3.11, 0.64       & \textbf{1.53, 0.24} & \textbf{1.98, 0.27} & \textbf{0.22} \\
room\_1   & 1.84, 0.34                           & 2.21, 0.42       & 0.21  & \textbf{1.83}, 0.35 & 2.22, 0.42       & 0.20              & 1.92, 0.35                           & 3.40, 0.74       & 1.96, \textbf{0.31} & \textbf{2.19, 0.38} & \textbf{0.17} \\
room\_2   & \textbf{1.31}, 0.26 & 2.69, 0.65       & 0.37  & 1.78, 0.29                           & 3.15, 0.69       & 0.39              & 2.97, 0.63                           & 13.0, 3.44       & 1.34, \textbf{0.22} & \textbf{2.57, 0.55} & \textbf{0.30} \\
office\_0 & 1.69, 0.34                           & 2.01, 0.45       & 0.17  & 1.79, 0.37                           & 2.24, 0.51       & 0.19              & \textbf{1.45, 0.30} & \textbf{1.92, 0.43}       & 1.61, 0.35                           & 1.97, 0.44                           & \textbf{0.15} \\
office\_1 & 2.10, 0.52                           & 2.23, 0.63       & 0.34  & 1.65, \textbf{0.42} & 1.99, 0.55       & 0.26              & 2.07, 0.53                           & 2.22, 0.59       & \textbf{1.54}, 0.44 & \textbf{1.74, 0.53} & \textbf{0.22} \\
office\_2 & 2.21, 0.41                           & 2.54, 0.49       & 0.29  & 2.07, 0.37                           & 2.28, 0.41       & 0.27              & 2.53, 0.51                           & 3.20, 0.64       & \textbf{1.69, 0.33} & \textbf{1.93, 0.37} & \textbf{0.25} \\
office\_3 & 2.13, 0.37                           & 4.91, 0.94       & 0.44  & \textbf{1.79, 0.28} & 5.72, 1.14       & 0.40              & 3.44, 0.63                           & 21.5, 7.95       & 2.40, 0.38                           & \textbf{4.25, 0.86} & \textbf{0.36} \\
office\_4 & 2.25, 0.43                           & 3.29, 1.03       & 0.40  & 2.42, 0.35                           & 3.57, 0.95       & 0.38              & 4.84, 0.90                           & 24.3, 6.18       & \textbf{1.69. 0.32} & \textbf{2.50, 0.86} & \textbf{0.35} \\ \hline
\end{tabular}
\label{tab:replica_result}
\vspace{-0.4cm}
\end{table*}


%

\subsection{Uncertainties of novel view synthesis}

Fig.~\ref{fig:uncertainty} qualitatively shows the error maps of the rendered RGB and depth images, as well as our predicted color and depth uncertainty maps. 
A comparison of the RGB and depth error maps reveals distinct distributions.
\Eg, in the example of the first row, large depth errors are mainly at structure borders, whereas large RGB errors mostly appear in highly textured areas. This observation verifies our idea of treating color and depth uncertainties separately. 
Further, a comparison of the error maps with the corresponding uncertainty maps ((c) to (d), (e) to (f)) shows a notable correlation, demonstrating the effectiveness of our estimated uncertainties. When using the rendered RGB-D data for other tasks, the uncertainty maps can provide valuable hints on the quality of the rendered data. 

\subsection{Camera pose estimation}
\label{sec:localize_results}

\noindent

\noindent
\textbf{Datasets.}
We evaluate our method on Replica~\cite{straub2019replica}, 12-Scenes~\cite{valentin2016learning}, and 7-Scenes~\cite{shotton2013scene} datasets.
Replica contains high-fidelity indoor scenes and is widely used by recent works of NeRFs and localization~\cite{zhi2021place,sucar2021imap,rosinol2022nerf,yang2022camera}. We use the sequences recorded in~\cite{zhi2021place},
choosing the first sequence of each scene for training and the second for testing. 
12-Scenes and 7-Scenes are both real-world indoor RGB-D datasets while the former has significantly larger environments. 
%


\noindent
\textbf{Evaluation.}
Instead of using thousands of training images, we conduct the few views training experiments. We first sample a small fraction of the original data to build $S_{real}$.
We create few-view datasets with a simple strategy: select one sequence from the training sequences of each scene and uniformly sample from it. 
For Replica, we uniformly sample 1/5 frames for the selected sequence.
For 12-Scenes, given that the sequences have varying numbers of frames, we adhere to uniform sampling but aim to achieve comparable sizes ($\sim$200, which is about 5\%$\sim$20\% frames for each scene) for all scenes as for other datasets.
%
%
For 7-Scenes, we choose one training sequence and uniformly sample 1/4 of its frames.
The numbers of selected novel views $S_{high}$ for Replica, 12-Scenes, and 7-Scenes are 150, 120, and 100, respectively. 

\noindent
\textbf{Baselines.}
We compare our method with SCRNet~\cite{li2020hierarchical}, SCRNet-ID~\cite{ng2021reassessing}, and SRC~\cite{dong2022visual}. SCRNet is the regression-only baseline proposed by~\cite{li2020hierarchical}, based on which we built SCRNet-ID and E-SCRNet. SCRNet-ID utilizes the \textit{In-Distribution} novel view selection policy proposed by~\cite{ng2021reassessing} to obtain new synthetic views. SRC~\cite{dong2022visual} is a recently proposed classification-based method for few-views scene-specific localization. We skip the experiments on 12-Scenes for SRC~\cite{dong2022visual} since they leveraged the dataset to pre-train the classification network with Reptile~\cite{nichol2018first} for model initialization.

\begin{table}[]
\setlength{\abovecaptionskip}{0.1cm}
\scriptsize
\setlength\tabcolsep{2pt}
\caption{Visual localization results on 7-Scenes and 12-Scenes. We report median translation and rotation errors (cm,\degree), and accuracy as the percentages of error \textless5cm, 5\degree. The best results are labeled in bold.}
\begin{tabular}{c|c|cc|cc|cc|cc}
\hline
\multicolumn{1}{l|}{} &            & \multicolumn{2}{c|}{SCRNet~\cite{li2020hierarchical}}                         & \multicolumn{2}{c|}{SCRNet-ID~\cite{ng2021reassessing}}                    & \multicolumn{2}{c|}{SRC~\cite{dong2022visual}}                                            & \multicolumn{2}{c}{Ours}                                            \\ \hline
\multicolumn{1}{l|}{} &            & Acc.$\uparrow$ & Med.$\downarrow$                   & Acc.$\uparrow$                 & Med.$\downarrow$ & Acc.$\uparrow$                 & Med.$\downarrow$                   & Acc.$\uparrow$                 & Med.$\downarrow$                   \\ \hline
\multirow{7}{*}{7S}   & chess      & 78.1           & 3.0, 1.1                           & 76.1                           & 3.1, 1.1         & 77.5                           & 3.6, 1.1                           & \textbf{80.2} & \textbf{2.7, 0.9} \\
                      & fire       & 75.9           & 3.4, 1.4                           & 74.1                           & 3.3, 1.3         & \textbf{96.0}                              & \textbf{1.7, 0.6}                                  & 85.3 & 2.6, 1.1 \\
                      & heads      & 97.8           & 1.4, \textbf{0.9} & 96.0                           & 1.4, 1.1         & \textbf{99.0} & 1.8, 1.2                           & 97.0                           & \textbf{1.3}, 1.0 \\
                      & office     & 59.0           & 4.3, 1.2                           & 45.2                           & 5.5, 1.5         & 42.3                           & 5.6, 1.4                           & \textbf{63.8} & \textbf{3.8, 1.1} \\
                      & pumpkin    & 44.9           & 5.4, 1.3                           & 43.0                           & 5.6, 1.3         & 42.0                           & 5.8, 1.5                           & \textbf{47.3} & \textbf{5.2, 1.3} \\
                      & redkitchen & 31.6           & 7.1, 2.0                           & 33.5                           & 7.0, 2.1         & 25.6                           & 6.9, \textbf{1.8} & \textbf{34.8} & \textbf{6.8}, 1.9 \\
                      & stairs     & 43.3           & 5.5, 1.5                           & 50.9                           & 4.9, 1.3         & 48.2                           & 5.1, 1.4                           & \textbf{61.3} & \textbf{4.5, 1.1} \\ \hline
\multirow{12}{*}{12S} & kitchen-1  & 90.4           & 2.3, 1.3                           & 87.1                           & 2.6, 1.4         & -                              & -                                  & \textbf{100}  & \textbf{0.9, 0.5} \\
                      & living-1   & 92.6           & 2.4, 0.8                           & 91.4                           & \textbf{2.0}, 0.8         & -                              & -                                  & \textbf{97.6} & 2.1, \textbf{0.6} \\
                      & Bed        & 73.5           & 3.3, 1.5                           & 82.3                           & 2.0, 0.8         & -                              & -                                  & \textbf{97.5} & \textbf{1.6, 0.7} \\
                      & kitchen-2  & 88.5           & 2.1, 1.0                           & 90.5                           & 1.8, 0.9         & -                              & -                                  & \textbf{97.1} & \textbf{1.2, 0.5} \\
                      & living-2   & 61.8           & 4.2, 1.8                           & 79.9                           & 3.0, 1.2         & -                              & -                                  & \textbf{95.1} & \textbf{2.0, 0.8} \\
                      & luke       & 58.6           & 4.4, 1.4                           & 73.3                           & 3.7, 1.3         & -                              & -                                  & \textbf{90.0} & \textbf{2.6, 1.0} \\
                      & gates362   & 88.6           & 2.6, 0.8                           & 87.6                           & 2.1, 1.0         & -                              & -                                  & \textbf{91.0} & \textbf{2.0, 0.8} \\
                      & gates381   & 76.3           & 3.4, 1.4                           & \textbf{82.8} & 2.9, 1.2         & -                              & -                                  & 81.4                           & \textbf{2.7, 1.2} \\
                      & lounge     & 86.9           & 2.7, 0.9                           & 78.9                           & 3.4, 1.1         & -                              & -                                  & \textbf{97.0} & \textbf{1.8, 0.6} \\
                      & manolis    & 84.0           & 1.8, 1.0                           & 85.3                           & 2.6, 1.2         & -                              & -                                  & \textbf{94.1} & \textbf{1.6, 0.7} \\
                      & 5a         & 65.9           & 3.6, 1.5                           & 72.8                           & 3.3, 1.2         & -                              & -                                  & \textbf{80.3} & \textbf{2.5, 0.9} \\
                      & 5b         & 64.7           & 3.4, 1.2                           & 66.7                           & 3.8, 1.3         & -                              & -                                  & \textbf{81.5} & \textbf{2.6, 0.8} \\ \hline 
\end{tabular}
\label{tab:7s_12s_result}
\vspace{-0.6cm}
\end{table}

\noindent
\textbf{Localization results.}
As shown in Table~\ref{tab:replica_result} and Table~\ref{tab:7s_12s_result}, across all three datasets, our method achieves the best performance compared with all baselines. Our method significantly outperforms SCRNet, showing that incorporating synthesized novel views could effectively enhance performance by a large margin. A comparison between ours and SCRNet-ID indicates that our novel view selection policy can select more informative samples. Additionally, compared with SRC, it can be seen that our method makes better use of the available data and boosts the localization performance more. 

\subsection{Analysis}
\label{sec:analysis}

\noindent
\textbf{Novel views selection.}
We compare different selection policies of novel views on the Replica dataset.
For each scene, we sort all the rendered views according to the epistemic uncertainty, then select the top 150 novel views with the highest score (\emph{High score} policy), 150 novel views with the lowest score (\emph{Low score} policy), and another 150 random novel views (\emph{Random} policy) for comparison. We use these two novel view sets to finetune the same network, respectively. From Table~\ref{tab:replica_policy}, we can see that the \emph{High score} policy significantly enhances the localization performance while the \emph{Low score} policy does not. It shows that the novel views with high scores provide high information gain to the model.
%

\begin{table}[t]
\setlength{\abovecaptionskip}{0.1cm}
\centering
\scriptsize
\setlength\tabcolsep{4.5pt}
\caption{Views selection policy. We report median and mean pose errors in (cm,\degree), and the sum of the L1 distance between predicted scene coordinates and ground truth in $m$.}
\begin{tabular}{ccccc}
\hline
\textbf{Scene} & \textbf{Policy} & \textbf{Median$\downarrow$}      & \textbf{Mean$\downarrow$}        & \textbf{Dist.$\downarrow$} \\ \hline
room\_0        & Low score       & 2.07, 0.315          & 2.15, 0.322          & 0.232          \\
               & Random          & 1.84, 0.290          & 2.24, 0.318          & 0.227          \\
               & High score      & \textbf{1.53, 0.243} & \textbf{1.98, 0.270} & \textbf{0.219} \\ \hline
room\_1        & Low score       & 2.17, 0.346          & 2.47, 0.418          & 0.191          \\
               & Random          & 2.32, 0.366          & 2.58, 0.417          & 0.186          \\
               & High score      & \textbf{1.96, 0.312} & \textbf{2.19, 0.389} & \textbf{0.173} \\ \hline
room\_2        & Low score       & 1.64, 0.291          & 3.04, 0.671          & 0.387          \\
               & Random          & 1.86, 0.253          & 2.97, 0.635          & 0.314          \\
               & High score      & \textbf{1.34, 0.220} & \textbf{2.57, 0.557} & \textbf{0.302} \\ \hline
\end{tabular}
\label{tab:replica_policy}
\vspace{-0.6cm}
\end{table}

\noindent
\textbf{Removing noisy regions with NeRF uncertainty.}
\label{sec:remove_noise}
As shown in Fig.~\ref{fig:pcd_filter}, the point cloud (left) generated from the raw RGB-D rendered by NeRF is noisy and might mislead the SCR model.
With uncertainty estimation from U-NeRF, we can identify the low-quality regions. By applying the pruning and weighting scheme mentioned in Sec.~\ref{sec:novel_views_selection}, we can eliminate unsatisfying parts from the raw output, resulting in a cleaner point cloud (compare middle vs. right).  
From Table~\ref{tab:scrnet_ue_nv}, we can also observe that removing noisy parts improves the performance of E-SCRNet, demonstrating the importance of applying pruning and weighting using U-NeRF uncertainty.


\begin{table}[h]
\vspace{-0.2cm}
\setlength{\abovecaptionskip}{0.1cm}
\centering
\scriptsize
\setlength\tabcolsep{4pt}
\caption{Effectiveness of evidential deep learning (EDL), adding novel views (NV), and U-NeRF uncertainty pruning. On scene \textit{luke} of 12-Scenes.}
\begin{tabular}{cccc}
\hline
\textbf{Method}  & \textbf{Acc.$\uparrow$} & \textbf{Median$\downarrow$} & \textbf{Mean$\downarrow$} \\ \hline
SCRNet           & 58                & 4.39, 1.438     & 5.46, 1.930   \\
\rotatebox[origin=c]{180}{$\Lsh$} + EDL      & 81                & 3.31, 1.289     & 3.73, 1.473   \\
\rotatebox[origin=c]{180}{$\Lsh$} + EDL + NV & 85                & 3.11, 1.115     & 3.36, 1.310   \\ 
\rotatebox[origin=c]{180}{$\Lsh$} + EDL + NV + pruning & \textbf{90}                & \textbf{2.61, 1.020}    & \textbf{2.89, 1.187}   \\ \hline
\end{tabular}
\label{tab:scrnet_ue_nv}
\vspace{-0.5cm}
\end{table}

\begin{figure}[ht]
\setlength{\abovecaptionskip}{0.1cm}
    \centering
     \begin{subfigure}[b]{0.13\textwidth}
         \centering
         \includegraphics[width=\textwidth]{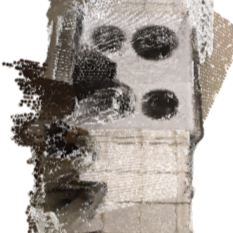}
         \caption{Raw}
     \end{subfigure}
     \begin{subfigure}[b]{0.13\textwidth}
         \centering
         \includegraphics[width=\textwidth]{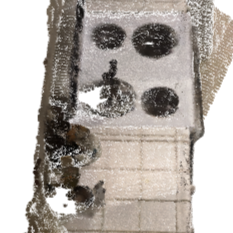}
         \caption{Pruned}
     \end{subfigure}
    \begin{subfigure}[b]{0.13\textwidth}
         \centering
         \includegraphics[width=\textwidth]{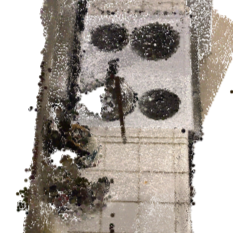}
         \caption{Ground truth}
     \end{subfigure}
    \caption[filter]{Removing noisy regions using NeRF uncertainty.}
    \label{fig:pcd_filter}
    \vspace{-0.35cm}
\end{figure}




\noindent
\textbf{Scene coordinate uncertainty.}
\label{sec:SCR_uncertainty}
To evaluate the effectiveness of the predicted scene coordinate uncertainty, we train the SCRNet and E-SCRNet on the whole real training set with the same number of epochs, in \textit{chess} of 7-Scenes. Then we compare their localization performance. In addition, for E-SCRNet, we sort the predicted 2D-3D correspondences with the estimated epistemic uncertainty. We then select the top 60\% correspondences with the lowest uncertainty (denoted as \textit{confident}) and the other correspondences (denoted as \textit{uncertain}), and perform RANSAC-PnP to estimate poses, respectively.
%
%
As shown in Table~\ref{tab:scr_uncertainty_pnp}, by comparing the \nth{1} and the \nth{2} row, E-SCRNet yields more accurate pose estimations than SCRNet when using all predicted correspondences. (It is also shown by comparing the \nth{1} and the \nth{2} row of Table~\ref{tab:scrnet_ue_nv}.)
E-SCRNet mitigates the influence of noisy samples and can better memorize the correct scene geometry during training, highlighting the importance of uncertainty estimation. Furthermore, the comparison between the \nth{3} and the \nth{4} row of Table~\ref{tab:scr_uncertainty_pnp} demonstrates that using correspondences with high confidence (low uncertainty) for camera pose optimization leads to more precise estimates. The correspondences with high uncertainty could noticeably downgrade the localization accuracy and increase pose errors, indicating the need to filter them out. It shows that the predicted uncertainty is meaningful. 
Lastly, if we compare the \nth{2} and the \nth{4} row of Table~\ref{tab:scr_uncertainty_pnp}, we can see that by using only the top 60\% of confident correspondences we achieve the same localization performance as using all correspondences. More importantly, by using fewer correspondences we also largely speed up the running time for RANSAC, which improves the overall inference efficiency and benefits robotic applications.

\begin{table}[t]
\setlength{\abovecaptionskip}{0.1cm}
\centering
\scriptsize
\setlength\tabcolsep{6pt}
\caption{Visual localization results on scene \textit{chess} of 7-Scenes. We report median translation and rotation errors (cm,\degree), accuracy as the percentages of error \textless5cm, 5\degree, and the run time of RANSAC-PnP during pose estimation for one query image in milliseconds. The best results are labeled in bold.}
\begin{tabular}{c|ccc}
\hline
                    & \textbf{Acc.$\uparrow$} & \textbf{Med.$\downarrow$}      & \textbf{Run time$\downarrow$} \\ \hline
SCRNet~\cite{li2020hierarchical}, all         & 95.4          & 2.4, 0.73          & 30                 \\
E-SCRNet, all       & \textbf{97.4} & \textbf{2.0, 0.71} & 30              \\
E-SCRNet, uncertain & 93.6          & 2.4, 0.84          & \textbf{11}                 \\
E-SCRNet, confident & \textbf{97.4} & \textbf{2.0, 0.71} & \textbf{11}     \\ \hline
\end{tabular}
\label{tab:scr_uncertainty_pnp}
\vspace{-0.7cm}
\end{table}

\noindent
\textbf{Comparing with all views results.} 
We mainly focus on data efficiency so the few-view setting used was particularly challenging.
As shown in Table~\ref{tab:additional_experiments}, with 50\% data our approach outperforms SCRNet trained on the full set and achieves even better performance when using all training data, indicating that our method requires only a small portion of training data but delivers comparable or even better performances than its counterpart trained on the full set.


\begin{table}[h]
\setlength{\abovecaptionskip}{0.1cm}
\vspace{-0.3cm}
\centering
\scriptsize
\setlength\tabcolsep{3pt}
\caption{Comparing with all views results. (Acc. / Med. pose errors).}
\begin{tabular}{c|cccc}
\hline
\textbf{12Scenes} & \textbf{SCRNet~\cite{li2020hierarchical} (All)} & \textbf{Ours (15\%)} & \textbf{Ours (50\%)}                                          & \textbf{Ours (All)}                                           \\ \hline
luke                               & 93.8 / 2.0, 0.9                        & 90.0 / 2.6, 1.0                       & 94.7 / 1.9, 0.7                                                                & \textbf{95.8 / 1.4, 0.6} \\
5b                                 & 93.3 / 1.9, 0.6                        & 81.5 / 2.6, 0.8                       & 95.8 / \textbf{1.7, 0.5} & \textbf{99.8 / 1.7, 0.5} \\ \hline
\end{tabular}
\label{tab:additional_experiments}
\vspace{-0.3cm}
\end{table}

\noindent
\textbf{Training time.}
To give an example, for \textit{manolis}, training SCRNet with full set takes $\sim$2 days. For our pipeline, training U-NeRF and rendering images takes $\sim$6h, training E-SCRNet $\sim$5h, finetuning it $\sim$5h, together $\sim$16h. Such improvement was consistently observed on all used datasets. Moreover, recent advances in NeRF, which achieve speedy optimization and rendering, could further reduce the training time.

\noindent
\textbf{Applying to other SCR systems.}
We also apply our approach with DSAC*~\cite{brachmann2021dsacstar}, improving its accuracy from 82.5\% to 92.8\%, and decreasing median translation and rotation errors from (3.3cm, 1.3\degree) to (2.0cm, 0.7\degree) on \textit{living-2} under the few views training setting, showing that our method can be applied to other SCR systems as a plug-in module, and enhance visual localization performances in a resource-efficient manner.
%

\section{Conclusion}
We present a novel pipeline that leverages NeRF to generate RGB-D pairs for training SCR networks. By modeling the uncertainty in NeRF, we are able to filter out noisy regions and artifacts in the rendered data. By formulating SCR from the evidential deep learning perspective, we model the uncertainty over the predicted scene coordinates. We proposed an uncertainty-guided novel view selection policy that could select the samples that bring the most information gain and promote localization performance with the highest efficiency. Our method could be beneficial to many robotic applications.




\clearpage

\bibliographystyle{./IEEEtran} 
\bibliography{./IEEEabrv.bib,./reference.bib}

\end{document}